\documentclass[letterpaper]{article} 
\usepackage{aaai2027}  
\nocopyright
\usepackage[hyphens]{url}  
\usepackage{graphicx} 
\usepackage{natbib}  
\usepackage{caption} 
\usepackage{algorithm}
\usepackage{algorithmic}

\usepackage{newfloat}
\usepackage{listings}
\DeclareCaptionStyle{ruled}{labelfont=normalfont,labelsep=colon,strut=off} 
\floatstyle{ruled}
\newfloat{listing}{tb}{lst}{}
\floatname{listing}{Listing}

\usepackage{booktabs}

\usepackage{multirow}
\usepackage{array}
\usepackage{amsmath,amssymb,amsfonts}
\usepackage{makecell}
\usepackage{pifont}
\newcommand{\cmark}{\ding{51}}
\newcommand{\xmark}{\ding{55}}
\title{LongEarth-R1: Benchmarking and Aligning Vision-Language Models for Long-Horizon Earth Observation Reasoning}

\author{
	Yupan Ding\textsuperscript{\rm 1},
	Jing Xiao\textsuperscript{\rm 1},
	Zhenyuan Zhang\textsuperscript{\rm 2},
	Chaofeng Chen\textsuperscript{\rm 1},\\
	Liang Liao\textsuperscript{\rm 3},
	Gui-Song Xia\textsuperscript{\rm 1},
	Mi Wang\textsuperscript{\rm 4}
}

\affiliations{
	\textsuperscript{\rm 1}School of Artificial Intelligence, Wuhan University, Wuhan, China\\
	\textsuperscript{\rm 2}School of Computer Science, Wuhan University, Wuhan, China\\
	\textsuperscript{\rm 3}Xi'an University of Electronic Science and Technology, Xi'an, China\\
	\textsuperscript{\rm 4}State Key Laboratory of Information Engineering in Surveying, Mapping and Remote Sensing, Wuhan University, Wuhan, China
}

\begin{document}
	
	\maketitle
	
	\begin{abstract}
		Long-horizon Earth observation reasoning requires models to organize multi-stage geographic evolution, localize spatial changes, detect temporal anomalies, and infer future from extended image sequences. However, existing remote sensing vision-language models mainly focus on isolated images, image pairs, or short sequences, limiting reliable grounding in the relevant frames and regions. We introduce LongEarth-Bench, a benchmark containing approximately 120k question-answering samples derived from 117k unique images. Its sequences average 15.14 frames and extend to 30 frames, covering 12 tasks across evolution summarization, spatial reasoning, anomaly identification, and logical prediction. A 30k-sample subset further provides structured reasoning traces linking key frames and changed regions to final answers. We develop LongEarth through supervised fine-tuning with explicit sequence identifiers and structured chain-of-thought supervision. Building on LongEarth, LongEarth-R1 applies group relative policy optimization with format, temporal, and spatial rewards. LongEarth-R1 achieves the best results on all 12 long-sequence tasks while remaining competitive on standard remote sensing benchmarks.
	\end{abstract}

	\section{Introduction}
	
	Remote sensing analysis is increasingly moving beyond isolated observations toward understanding how geographic regions evolve over time~\cite{weng2025vlmrsurvey}. With the growing availability of high-revisit Earth observation imagery, it is now possible to observe multi-stage processes such as urban construction, disaster evolution, and ecosystem recovery~\cite{liu2024temporalvlmsurvey,irvin2025teochat,jain2026timesenclip}. For example, given observations before, during, and after a flood, a model should identify its onset and affected regions, track its evolution, and infer the recovery stage. This requires more than pairwise change detection: the model must recover a temporal trajectory, localize supporting frames and regions, and separate geographic evolution from seasonal and acquisition-induced variations~\cite{li2024unirs,soni2025earthdial,luo2026vlrsbench}. We refer to this capability as \textit{long-horizon Earth observation reasoning}: reasoning about multi-stage geographic evolution from extended Earth observation sequences.
	
	Recent remote sensing vision-language models (RSVLMs) have progressively extended language-guided interpretation from individual observations to image pairs and temporal sequences. Single-temporal models connect an individual observation with language through image description, visual question answering, and visual grounding~\cite{zhang2024earthgpt,hu2025rsgpt}. Bi-temporal models further compare paired observations to describe and localize geographic changes~\cite{yang2024kcfi,liu2024changeagent}. More recent temporal RSVLMs accept multiple observations and support sequence-level dialogue or change understanding~\cite{irvin2025teochat,soni2025earthdial,xuan2025dynamicvl}. Nevertheless, these capabilities are still insufficient for long-horizon reasoning. A single image contains no evolution trajectory, an image pair exposes mainly the endpoints of change, and current sequence models generally emphasize recognition or descriptive question answering rather than reconstructing a complete multi-stage process. They consequently remain limited in tracing critical transitions, grounding conclusions across frames and regions, and inferring how an evolving process may continue.
	
	\begin{figure*}[tb]
		\centering
		\includegraphics[width=\textwidth]{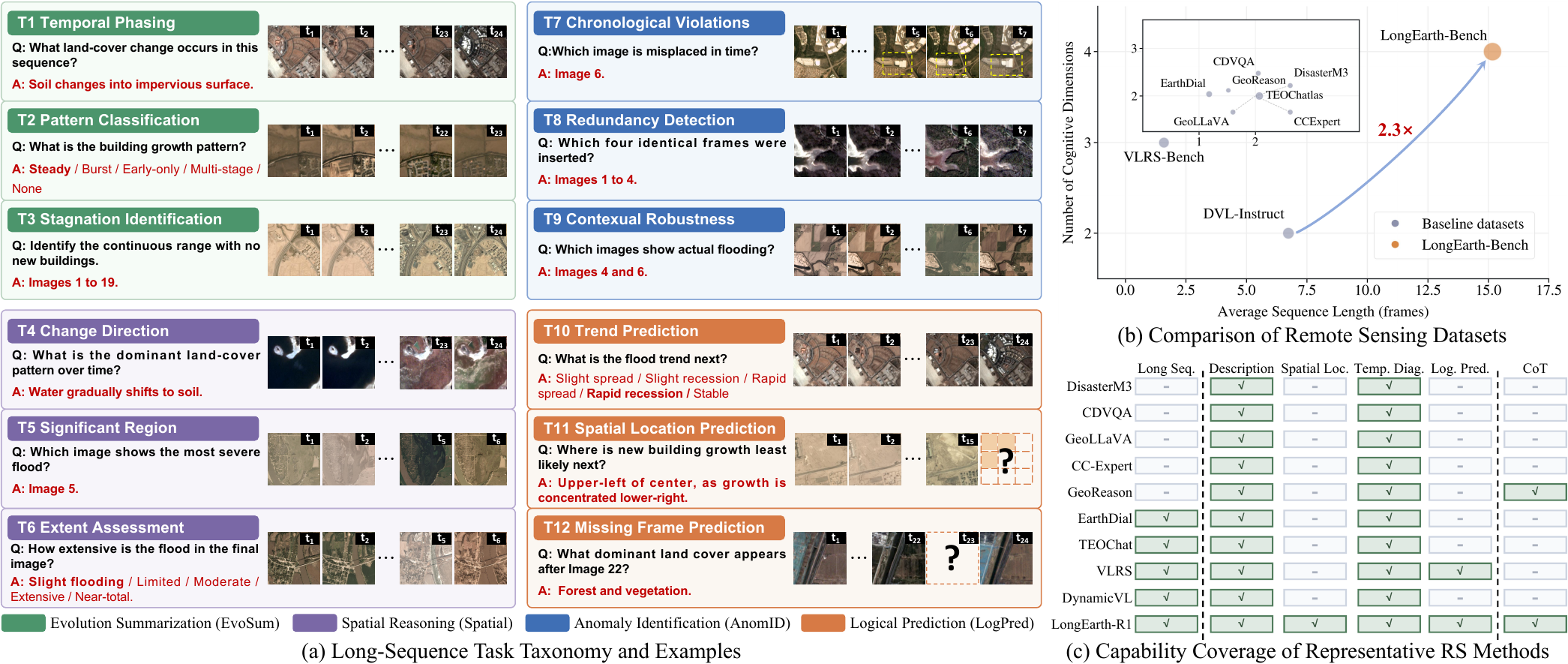}
		\vspace{-6mm}
		\caption{\small
			Task taxonomy, benchmark positioning, and model capability comparison of LongEarth-Bench and LongEarth-R1.
			(a) Representative examples of the 12 tasks grouped into four cognitive dimensions.
			(b) Average sequence length versus cognitive-dimension coverage; bubble size denotes maximum sequence length.
			(c) Capability coverage of representative remote sensing methods, including long-sequence understanding, evolution description, spatial localization, temporal diagnosis, logical prediction, and CoT.
		}
		\vspace{-5mm}
		\label{fig:longearth_overview}
	\end{figure*}
	
	To address these limitations, we formulate long-horizon Earth observation reasoning as reasoning over extended sequences. This capability requires models to identify how geographic entities evolve, localize and characterize their changes, assess whether the observed evolution is temporally consistent, and infer unobserved states. We therefore organize it into four progressive cognitive dimensions: 1) \textit{Evolution summarization} organizes major stages into coherent temporal trajectories; 2) \textit{Spatial reasoning} localizes changes and models their evolving spatial relations; 3) \textit{Anomaly identification} detects chronological violations, repetitions, and contextual interference; and 4) \textit{Logical prediction} infers subsequent or missing states from accumulated evidence. Figure~\ref{fig:longearth_overview}(a) illustrates this hierarchy through 12 fine-grained tasks.
	
	Based on the four cognitive dimensions, we construct \textbf{LongEarth-Bench}, a large-scale benchmark for long-horizon remote sensing spatiotemporal reasoning. It contains approximately 120k question-answering (QA) samples derived from 117k images, with an average sequence length of 15.14 frames. The benchmark covers diverse geographic regions, land-cover categories, temporal spans, change rates, and multi-stage processes. Its QA samples are generated from segmentation annotations, geometric relations, and temporal change trajectories. Moreover, 30k samples contain structured reasoning annotations connecting frame selection, temporal localization, spatial evidence, and final conclusions. Figure~\ref{fig:longearth_overview}(b) compares LongEarth-Bench with existing benchmarks according to average sequence length and coverage of the four cognitive dimensions.
	
	To enable the RSVLMs with long-horizon spatiotemporal reasoning, we first develop \textbf{LongEarth} through supervised fine-tuning with explicit sequence identifiers and structured chain-of-thought (CoT) traces that connect key frames, temporal changes, and changed regions to final answers. The former establishes stable frame-level temporal anchors, while the latter teaches the model to select relevant observations and integrate temporal and spatial evidence across the sequence. Building on LongEarth, \textbf{LongEarth-R1} applies group relative policy optimization (GRPO) with complementary format, temporal, and spatial rewards to optimize output completeness, evolution-stage localization, temporal ordering, key-frame selection, and changed-region consistency beyond final-answer correctness. Together, LongEarth and LongEarth-R1 establish reproducible supervised and reinforcement-learning baselines on LongEarth-Bench. Figure~\ref{fig:longearth_overview}(c) compares LongEarth-R1 with representative remote sensing methods across six model capabilities, with LongEarth-R1 covering all six.
	
	The main contributions of this paper are as follows.
	
	\begin{itemize}
		\item We formulate long-horizon Earth observation reasoning as reasoning over multi-stage geographic evolution through four cognitive dimensions covering evolution summarization, spatial reasoning, anomaly identification, and logical prediction.
		
		\item We introduce LongEarth-Bench, a large-scale benchmark with approximately 120k samples. It covers diverse geographic processes and includes 30k samples with structured annotations for evidence-grounded reasoning.
		
		\item We develop LongEarth through supervised fine-tuning and LongEarth-R1 through GRPO-based temporal and spatial rewards, and systematically evaluate current RSVLMs across sequence lengths, evolution processes, and reasoning dimensions.
	\end{itemize}

	\section{Related Work}
	
	\begin{figure*}[tb]
		\centering
		\includegraphics[width=0.97\textwidth]{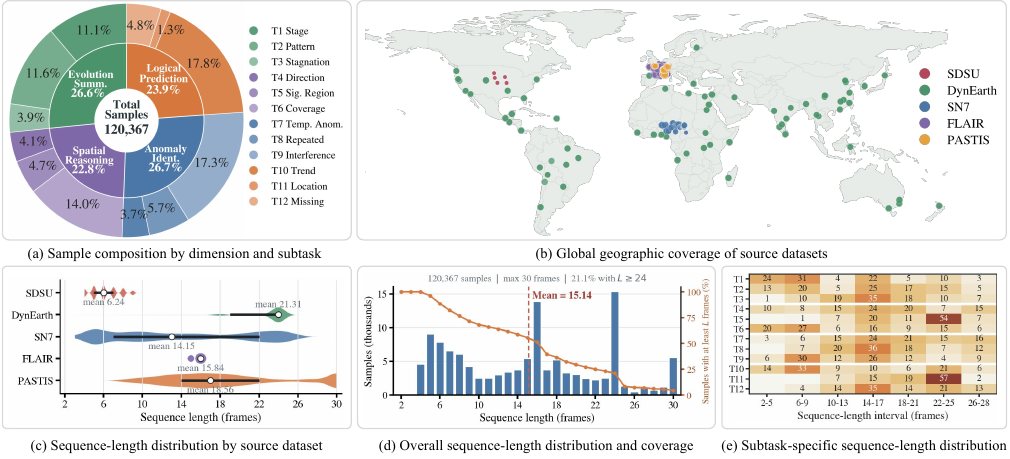}
		\vspace{-2mm}
		\caption{\small
			LongEarth-Bench composition, geographic coverage, and sequence-length statistics.
			(a) Sample distribution across four cognitive dimensions (inner ring) and 12 tasks (outer ring).
			(b) Geographic coverage; colors denote the five source datasets and marker size the local mean sequence length.
			(c) Source-specific length distributions with annotated means.
			(d) Overall length distribution; bars show counts and the curve gives the fraction of sequences at least each length.
			(e) Task-conditioned length, with rows normalized within each subtask.
		}
		\vspace{-5mm}
		\label{fig:longearth_analysis}
	\end{figure*}

	\subsection{Single- to Multi-Temporal RSVLMs}
	
	RSVLMs have progressed from static image-language alignment to temporal Earth observation understanding. Early models such as RemoteCLIP~\cite{liu2023remoteclip}, GeoRSCLIP~\cite{zhang2023rs5m}, GeoChat~\cite{kuckreja2024geochat}, and EarthGPT~\cite{zhang2024earthgpt} focus on aligning a single observation with language. VHM~\cite{pang2025vhm} extends this capability to scene classification, visual question answering, and visual grounding, while EarthVQA~\cite{wang2024earthvqa} advances relational visual question answering and spatial reasoning. Beyond remote sensing, SpaceVLLM~\cite{wang2026spacevllm} studies frame-specific spatiotemporal video grounding. RSICCformer~\cite{liu2022rsiccformer}, GeoLLaVA~\cite{elgendy2024geollava}, Change-Agent~\cite{liu2024changeagent}, and CCExpert~\cite{wang2024ccexpert} extend language-based analysis to bi-temporal image pairs through change captioning, detection, or difference-aware integration. 
	DisasterM3~\cite{wang2025disasterm3} provides a bi-temporal remote sensing vision-language dataset and benchmark for disaster assessment and response. However, existing bi-temporal settings cannot explicitly represent the intermediate stages of an evolving geographic process.
	
	Recent multi-temporal RSVLMs process extended Earth observation sequences. EarthDial~\cite{soni2025earthdial} supports multispectral, multi-temporal, and multi-resolution conversational inputs, TEOChat~\cite{irvin2025teochat} enables dialogue and question answering over temporal observations, using sequences that average 2.07 frames and extend to at most 8 frames, and UniRS~\cite{li2024unirs} unifies single-image, bi-temporal, and video tasks. TimeSenCLIP~\cite{jain2026timesenclip} learns spectral-temporal representations from Sentinel-2 time series, while DynamicVL~\cite{xuan2025dynamicvl} benchmarks dynamic city understanding with multi-temporal scenes averaging 6.73 frames and extending to at most 10 frames, while VLRS-Bench~\cite{luo2026vlrsbench} evaluates cognition, decision, and prediction with sequences averaging 1.59 frames and covering up to eight temporal phases. These studies establish important foundations for multi-temporal remote sensing understanding. However, extended trajectories spanning many evolution stages remain less studied, particularly when evaluation requires anomaly localization, cross-stage spatial reasoning, and identification of the observations that support a conclusion.

	\subsection{Reinforcement Learning for Reasoning in RSVLMs}
	
	Recent RSVLMs have incorporated structured reasoning and reinforcement learning to support more interpretable and evidence-grounded geospatial analysis. Multimodal-CoT separates rationale generation from answer inference~\cite{zhang2023multimodalcot}, while Geo-CoT constructs perceptually grounded reasoning traces and trains RSThinker through supervised fine-tuning followed by GRPO~\cite{liu2025geocot}. Beyond remote sensing, IAD-R1~\cite{li2026iadr1} similarly combines CoT-based supervised fine-tuning with structured GRPO for consistent vision-language reasoning. SAMChat~\cite{koksal2025milchat} adopts CoT supervision and GRPO for small-scale remote sensing analysis, while RemoteReasoner~\cite{yao2026remotereasoner} applies reinforcement learning to a unified workflow covering object-, region-, and pixel-level geospatial reasoning. GeoReason~\cite{li2026georeason} aligns reasoning and answers through logical-consistency reinforcement learning, while GeoChain~\cite{yerramilli2025geochain} studies multimodal CoT geographic reasoning. In parallel, VLRS-Bench evaluates complex reasoning over single-temporal and multi-temporal remote sensing inputs~\cite{luo2026vlrsbench}.
	
	These studies show the value of intermediate supervision and task-specific rewards for reasoning beyond final-answer prediction. 
	However, existing methods primarily reason over individual observations or restricted temporal settings, without explicitly aligning intermediate reasoning with evidence distributed across long Earth observation sequences.

	\section{The LongEarth-Bench Dataset}
	
	LongEarth-Bench defines this hierarchy through 12 fine-grained tasks. We abbreviate the four cognitive dimensions as evolution summarization (EvoSum), spatial reasoning (Spatial), anomaly identification (AnomID), and logical prediction (LogPred). Figure~\ref{fig:longearth_analysis}(a) reports the sample distribution. The 120,367 samples are broadly balanced across EvoSum (26.6\%), Spatial (22.8\%), AnomID (26.7\%), and LogPred (23.9\%), while preserving diversity across the 12 tasks.
	
	\subsection{Dataset Construction}
	
\begin{figure}[tb]
		\centering
		\includegraphics[width=0.99\columnwidth]{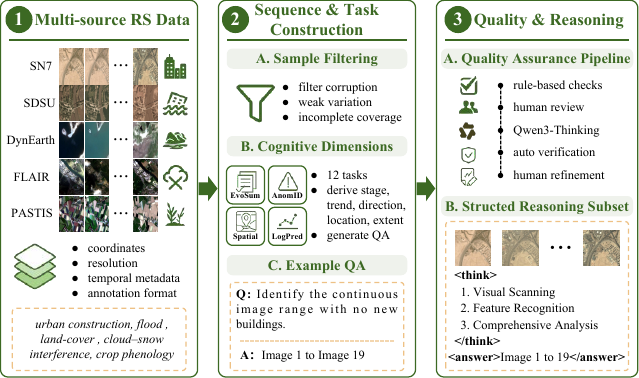} 
		\vspace{-1mm}
		\caption{\small
			Construction pipeline of LongEarth-Bench.
		}
		\label{fig:longearth_construction}
		\vspace{-5mm}
	\end{figure}

	Figure~\ref{fig:longearth_construction} presents the construction pipeline of LongEarth-Bench: multi-source data integration, sequence filtering and cognitive task construction, followed by quality-controlled structured reasoning annotation. Specifically, LongEarth-Bench is derived from SpaceNet 7 (SN7)~\cite{vanetten2021spacenet7}, SDSU MidWest Flood (SDSU)~\cite{jang2024sdsuflood}, DynamicEarthNet (DynEarth)~\cite{toker2022dynamicearthnet}, FLAIR\#2 (FLAIR)~\cite{garioud2023flair2}, and PASTIS-R (PASTIS)~\cite{garnot2022pastisr}. These sources cover urban construction, flood dynamics, natural land-cover evolution, cloud--snow interference, and crop phenology. As shown in Figure~\ref{fig:longearth_analysis}(b), they span North America, South America, Europe, Africa, Asia, and Oceania, reducing dependence on a single geographic region or evolution process. Coordinate systems are normalized, while spatial resolutions, temporal metadata, and annotation formats are standardized to construct temporally ordered and spatially aligned remote sensing image sequences. Samples with severe observation corruption, incomplete temporal coverage, registration failures, or insufficient long-term variation are filtered.
	
	Task annotations are generated from temporal trajectories together with segmentation, polygon, land-cover, and image-level flood evidence. Stage boundaries, trends, locations, directions, extents, observation quality, and phenological states are derived from cross-frame changes. Controlled order perturbations, repeated-frame insertions, and contextual disturbances are used to construct anomaly tasks, while prediction tasks are formed from historical trajectories and adjacent temporal states. Rule-based validation verifies sequence integrity, answer consistency, frame indices, spatial labels, and image paths, followed by human review for visual support and ambiguity.
	
	
	To complement answer-level supervision, we construct a balanced subset of 30k structured reasoning samples across the 12 tasks. Initial traces are generated by Qwen3-VL-8B-Thinking~\cite{bai2025qwen3vl} in a unified format: the \texttt{\textless think\textgreater} field contains visual scanning, feature identification, and integrated analysis, and the \texttt{\textless answer\textgreater} field retains the reference answer. 
	We automatically verify tag completeness, stage ordering, non-empty reasoning fields, and answer consistency; invalid outputs are regenerated. Human experts then assess visual grounding, chain coherence, and potential answer leakage. The resulting subset provides reliable process-level supervision for learning temporal stages, spatial dynamics, anomalies, and logical prediction.
	
	\subsection{Statistical Analysis of LongEarth-Bench}
	
	Figure~\ref{fig:longearth_analysis}(c) represents complementary temporal regimes across the five data sources. SDSU provides the shortest sequences, with a mean length of 6.24 frames, whereas DynEarth provides the longest, averaging 21.31 frames. SN7, FLAIR, and PASTIS cover intermediate and long contexts, with mean lengths of 14.15, 15.84, and 18.56 frames, respectively. These source-specific distributions preserve the temporal characteristics of disaster events, urban construction, natural-surface evolution, and crop phenology.
	
	As shown in Figure~\ref{fig:longearth_analysis}(d), LongEarth-Bench contains 120,367 samples in total with an average sequence length of 15.14 frames, a median of 16 frames, and a maximum of 30 frames. Compared with TEOChatlas~\cite{irvin2025teochat}, DVL-Instruct~\cite{xuan2025dynamicvl}, and VLRS-Bench~\cite{luo2026vlrsbench}, LongEarth-Bench is 7.3$\times$, 2.2$\times$, and 9.5$\times$ longer on average, and supports maximum sequences that are 3.75$\times$, 3.0$\times$, and 3.75$\times$ longer, respectively. Moreover, 21.1\% of the samples contain at least 24 observations. LongEarth-Bench therefore covers a broad spectrum from compact event sequences to extended multi-stage trajectories, providing a controlled setting for evaluating sequence-grounded long-horizon reasoning.
	
	Figure~\ref{fig:longearth_analysis}(e) summarizes the sequence-length distribution for each task of LongEarth-Bench and shows that all 12 tasks cover multiple sequence-length intervals, while their dominant temporal horizons differ according to their evidence requirements. Change-direction reasoning and spatial-location prediction rely on extended observations, with 53\% and 57\% of their samples falling within 22--25 frames, respectively. In contrast, contextual robustness and extent assessment contain more compact event-centered sequences. The substantial overlap across tasks also prevents sequence length from serving as a simple task shortcut.
	
	\section{Method}
	
	\begin{figure*}[tb]
		\centering
		\includegraphics[width=0.95\textwidth]{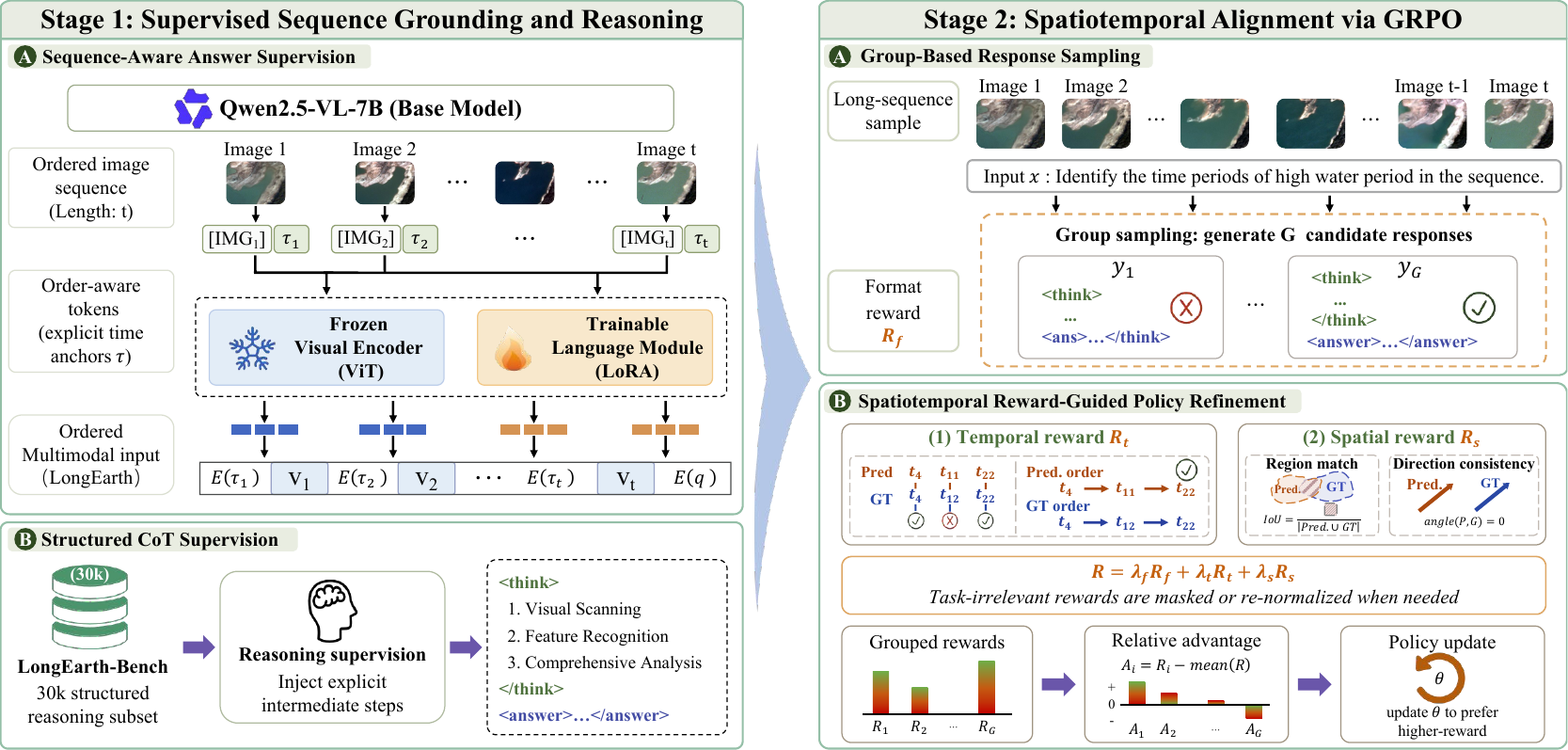} 
		\vspace{-2mm}
		\caption{\small
			Overview of LongEarth and LongEarth-R1. Stage 1 trains LongEarth through sequence-aware supervised fine-tuning with explicit sequence identifiers (Seq. IDs) and structured reasoning traces from the 30k-sample subset. Stage 2 initializes from LongEarth and applies GRPO to obtain LongEarth-R1 with format, temporal, and spatial rewards.
		}
		\label{fig:longearth_framework}
		\vspace{-4mm}
	\end{figure*}
	
	This section presents a two-stage framework for long-horizon spatiotemporal reasoning. As shown in Figure~\ref{fig:longearth_framework}, the framework is built on Qwen2.5-VL-7B~\cite{bai2025qwen25vl}. Stage 1 applies supervised fine-tuning (SFT) with two complementary forms of supervision. Sequence-aware answer supervision uses explicit sequence identifiers to establish stable frame-level temporal anchors, while structured CoT supervision teaches the model to select relevant observations and integrate temporal and spatial evidence across the sequence, yielding LongEarth. Stage 2 initializes from LongEarth and applies GRPO with complementary format, temporal, and spatial rewards to improve reasoning structure and spatiotemporal consistency, yielding LongEarth-R1.
	
	\subsection{Supervised Sequence Grounding and Reasoning}
	
	The first stage contains two supervised components. The first component performs sequence-aware answer supervision, which adapts the base model to long-term remote sensing inputs and establishes frame-level temporal anchors. The second component injects structured reasoning traces, which further guides the model to organize cross-frame evidence before producing the final answer.
	
	\paragraph{Sequence-Aware Answer Supervision.}
	
	This component adapts the base model to long-term remote sensing data and establishes the basic mapping from image sequences and questions to answers. Given a remote sensing sequence $X=\{I_1,\ldots,I_T\}$ with $T$ temporal observations and a question $q$, the model is required to generate an answer $a$ based on the full sequence. Since long-term tasks require both single-frame recognition and cross-frame change understanding, we assign each image an explicit sequence identifier (Seq. ID), such as Image 1, Image 2, $\ldots$, Image $T$.
	
	Let $V_t=f_v(I_t)$ denote the visual tokens of the $t$-th frame encoded by the visual encoder $f_v$. The ordered multimodal input is represented as:
	\begin{equation}
		\label{eq:ordered_input}
		H=[q;\mathrm{Image}\ 1,V_1;\ldots;\mathrm{Image}\ T,V_T].
	\end{equation}
	This representation provides stable frame-level anchors and reduces ambiguity in key-frame reference, temporal interval localization, and cross-frame relation modeling. During training, the visual encoder is frozen, while low-rank adaptation (LoRA) modules are applied to the language model layers for parameter-efficient alignment~\cite{hu2022lora}. For answer-only samples, this objective mainly constrains the final answer and does not explicitly supervise how the model selects cross-frame evidence or organizes intermediate reasoning.
	
	\paragraph{Structured CoT Supervision.}
	
	Within the supervised stage, we use the structured reasoning subset of LongEarth-Bench to supervise evidence-grounded responses. For each reasoning sample $(X,q,r,a)$, the target response contains a reasoning trace $r$ and final answer $a$, formatted as \texttt{\textless think\textgreater} $r$ \texttt{\textless /think\textgreater} \texttt{\textless answer\textgreater} $a$ \texttt{\textless /answer\textgreater}. 
	
	The reasoning trace follows three steps: visual scanning, feature identification, and integrated analysis. Visual scanning describes the main land-cover states and spatial layouts across temporal observations. Feature identification extracts key frames, changed regions, directions, or anomalies related to the question. Integrated analysis combines cross-frame evidence and produces the final conclusion. This design shifts the model from direct answer learning to process-level supervision. Specifically, given the target sequence $y=(y_1,\ldots,y_N)$, the structured-reasoning objective is:
	\begin{equation}
		\label{eq:cot_loss}
		\mathcal{L}_{\mathrm{CoT}}
		=
		-\sum_{n=1}^{N}
		\log p_{\theta}(y_n\mid y_{<n},H).
	\end{equation}
	This objective jointly supervises the reasoning trace and final answer, but reference-trace imitation alone cannot directly penalize temporal ordering errors or spatial inconsistencies. The second stage therefore applies GRPO to further optimize structural validity and spatiotemporal grounding.

	\begin{table*}[tb]
		\centering
		\vspace{-2mm}
		{
			\fontsize{7.6bp}{9.2bp}\selectfont
			
			\setlength{\tabcolsep}{0pt}
			\def\tableedge{6pt}            
			\def\categorytaskgap{3pt}      
			\def\taskrulegap{3pt}          
			\def\taskmodelgap{3pt}         
			\def\verticalrulewidth{0.4pt}  
			
			\renewcommand{\arraystretch}{1.05}
			\setlength{\aboverulesep}{0.25ex}
			\setlength{\belowrulesep}{0.25ex}
			
			\begin{tabular*}{0.98\textwidth}{
					@{\hspace{\tableedge}}
					>{\raggedright\arraybackslash}m{1.70cm}
					@{\hspace{\categorytaskgap}}
					!{\vrule width \verticalrulewidth}
					@{\hspace{\taskrulegap}}
					>{\raggedright\arraybackslash}m{3.45cm}
					@{\hspace{\taskmodelgap}\extracolsep{\fill}}
					>{\centering\arraybackslash}m{1.60cm}
					>{\centering\arraybackslash}m{1.10cm}
					>{\centering\arraybackslash}m{1.90cm}
					>{\centering\arraybackslash}m{1.15cm}
					>{\centering\arraybackslash}m{1.15cm}
					>{\centering\arraybackslash}m{1.20cm}
					>{\centering\arraybackslash}m{1.30cm}
					>{\centering\arraybackslash}m{1.90cm}
					@{\hspace{\tableedge}}
				}
				\toprule
				Category & Task & Video-LLaVA & Qwen2.5 & Qwen3-Thinking & TEOChat & TEOChat* & EarthDial & EarthDial* & \textbf{LongEarth-R1} \\
				\midrule
				\multirow[c]{3}{1.70cm}{\makecell[l]{Evolution\\Summarization}} & \mbox{T1 Temporal Phasing} & 24.53 & 41.22 & 49.44 & 32.02 & 64.16 & 31.84 & 46.18 & \textbf{77.31} \\
				& \mbox{T2 Pattern Classification} & 39.32 & 46.44 & 59.60 & 35.81 & 49.58 & 45.19 & 69.90 & \textbf{83.14} \\
				& \mbox{T3 Stagnation Identification} & 10.58 & 34.11 & 29.73 & 35.98 & 50.74 & 25.92 & 46.95 & \textbf{63.57} \\
				\midrule
				\multirow[c]{3}{1.70cm}{\makecell[l]{Spatial\\Reasoning}} & \mbox{T4 Change Direction} & 32.72 & 43.86 & 40.08 & 31.92 & 40.28 & 33.44 & 45.80 & \textbf{53.37} \\
				& \mbox{T5 Significant Region} & 34.65 & 40.53 & 36.47 & 31.60 & 50.45 & 32.39 & 62.69 & \textbf{68.96} \\
				& \mbox{T6 Extent Assessment} & 30.76 & 39.24 & 33.48 & 35.03 & 58.48 & 37.04 & 67.58 & \textbf{74.32} \\
				\midrule
				\multirow[c]{3}{1.70cm}{\makecell[l]{Anomaly\\Identification}} & \mbox{T7 Chronological Violations} & 2.86 & 12.38 & 7.87 & 7.90 & 15.47 & 5.37 & 15.92 & \textbf{50.80} \\
				& \mbox{T8 Redundancy Detection} & 5.90 & 39.78 & 36.68 & 10.31 & 27.57 & 14.51 & 20.32 & \textbf{90.28} \\
				& \mbox{T9 Contextual Robustness} & 18.60 & 35.65 & 37.45 & 15.51 & 48.74 & 12.42 & 49.12 & \textbf{89.12} \\
				\midrule
				\multirow[c]{3}{1.70cm}{\makecell[l]{Logical\\Prediction}} & \mbox{T10 Trend Prediction} & 29.78 & 46.78 & 32.03 & 36.30 & 55.44 & 20.61 & 60.54 & \textbf{76.42} \\
				& \mbox{T11 Spatial Location Prediction} & 47.60 & 50.61 & 33.77 & 28.48 & 55.98 & 49.85 & 87.62 & \textbf{89.87} \\
				& \mbox{T12 Missing Frame Prediction} & 29.69 & 33.83 & 47.86 & 38.40 & 58.94 & 36.19 & 62.04 & \textbf{64.25} \\
				\bottomrule
			\end{tabular*}
		}
		\caption{\small Performance comparison on the 12 long-sequence cognitive tasks in LongEarth-Bench (\%). An asterisk (*) denotes supervised fine-tuning on LongEarth-Bench. The best result in each row is shown in bold.}
		\label{tab:longearth_main}
		\vspace{-4mm}
	\end{table*}
	
	\subsection{Spatiotemporal Alignment via GRPO}
	
	Starting from LongEarth, the second stage applies GRPO to obtain LongEarth-R1 and align reasoning with long-term remote sensing objectives~\cite{shao2024deepseekmath,shen2025vlmr1}. Unlike SFT, which fits a single reference output, GRPO compares a group of candidate responses for the same input and updates the policy using relative rewards. It therefore improves answer quality together with reasoning structure and spatiotemporal consistency.
	
	For multimodal input $H$, the old policy $\pi_{\theta_{\mathrm{old}}}$ samples $G$ responses $y_i=(r_i,a_i)$, each comprising a reasoning trace and final answer. We assign each response a weighted reward:
	\begin{equation}
		\label{eq:total_reward}
		R_i=\lambda_f R_i^{\mathrm{fmt}}+\lambda_t R_i^{\mathrm{time}}+\lambda_s R_i^{\mathrm{space}},
	\end{equation}
	where the three terms measure format validity, temporal grounding, and spatial consistency, respectively. The format reward checks the required reasoning--answer structure and task-specific answer parsability. The temporal reward combines matching of referenced frames with their chronological consistency, while the spatial reward evaluates agreement on changed locations, directions, and extents. For tasks without temporal or spatial labels, inapplicable terms are omitted and the remaining weights are renormalized; detailed reward definitions are provided in the supplementary material.
	
	GRPO normalizes rewards within the response group as $A_i=(R_i-\mu_R)/(\sigma_R+\epsilon)$ and uses the policy ratio $\rho_i=\pi_\theta(y_i\mid H)/\pi_{\theta_{\mathrm{old}}}(y_i\mid H)$. Let $\bar{\rho}_i=\operatorname{clip}(\rho_i,1-\varepsilon,1+\varepsilon)$. The clipped objective is
	\begin{equation}
		\label{eq:grpo_objective}
		\mathcal{J}_{\mathrm{GRPO}}
		=\frac{1}{G}\sum_{i=1}^{G}\min\!\left(\rho_i A_i,\bar{\rho}_i A_i\right)-\beta D_{\mathrm{KL}}\!\left(\pi_\theta\Vert\pi_{\mathrm{ref}}\right),
	\end{equation}
	where $\varepsilon$ is the clipping coefficient, $\beta$ is the Kullback--Leibler (KL) weight, and $\pi_{\mathrm{ref}}$ is the policy obtained after the supervised stage. The KL term preserves the visual-language ability acquired during SFT. Together, the rewards shift learning from answer imitation toward structured, temporally ordered, and spatially grounded reasoning.

	\section{Experiments}
	
	\begin{table*}[tb]
		\centering
		\vspace{-2mm}
		{
			\fontsize{7.6bp}{9.0bp}\selectfont
			\setlength{\tabcolsep}{0pt}
			\def\tableedge{10pt}         
			\def\inputdatasetgap{4pt}    
			\def\datasetmodelgap{4pt}    
			\def\humanwidth{0.85cm}
			\def\videowidth{1.60cm}
			\def\geochatwidth{1.05cm}
			\def\qwenwidth{1.15cm}
			\def\teochatwidth{1.10cm}
			\def\earthwidth{1.25cm}
			\def\longearthwidth{1.70cm}
			
			\renewcommand{\arraystretch}{1.05}
			\setlength{\aboverulesep}{0.25ex}
			\setlength{\belowrulesep}{0.25ex}
			
			\begin{tabular*}{0.95\textwidth}{
					@{\hspace{\tableedge}}
					>{\centering\arraybackslash}m{1.45cm}
					@{\hspace{\inputdatasetgap}}
					>{\raggedright\arraybackslash}m{2.20cm}
					!{\vrule width 0.4pt}
					@{\hspace{\datasetmodelgap}}
					>{\centering\arraybackslash}m{\humanwidth}
					@{\extracolsep{\fill}}
					>{\centering\arraybackslash}m{\videowidth}
					>{\centering\arraybackslash}m{\geochatwidth}
					>{\centering\arraybackslash}m{\qwenwidth}
					>{\centering\arraybackslash}m{\earthwidth}
					>{\centering\arraybackslash}m{\teochatwidth}
					>{\centering\arraybackslash}m{\longearthwidth}
					@{\hspace{\tableedge}}}
				\toprule
				Category & Dataset & Human & Video-LLaVA & GeoChat & Qwen2.5 & EarthDial & TEOChat & \textbf{LongEarth-R1} \\
				\midrule
				\multirow{2}{*}{\makecell[l]{Single image}} & AID & -- & 52.40 & 72.00 & 65.60 & \textbf{88.67} & 80.90 & 86.03 \\
				& UCM & -- & 46.50 & 84.40 & 70.70 & 80.48 & 86.30 & \textbf{90.19} \\
				\midrule
				\multirow{4}{*}{Bi-temporal} & ABCD-CD & 95.20 & 50.00 & -- & 69.80 & 63.46 & 85.60 & \textbf{91.20} \\
				& CDVQA-QA & 63.40 & 29.80 & -- & 50.50 & 47.47 & 47.20 & \textbf{56.60} \\
				& xBD (Avg.) & 56.20 & 21.20 & 25.10 & 36.80 & 24.78 & 59.60 & \textbf{71.50} \\
				& S2Looking (Avg.) & 26.50 & 24.80 & 32.70 & 8.20 & 29.70 & 57.70 & \textbf{60.50} \\
				\midrule
				\multirow{3}{*}{\makecell[c]{Multi-image\\$(\leq 8)$}} & Qfabric (Avg.) & 71.00 & 17.60 & 18.40 & 23.70 & 33.59 & \textbf{70.80} & 69.70 \\
				& fMoW-LR-TSC & -- & 4.90 & 26.30 & 0.10 & 24.08 & 45.50 & \textbf{45.60} \\
				& fMoW-HR-TSC & 65.90 & 16.60 & 59.20 & 28.50 & 60.86 & \textbf{75.10} & 72.90 \\
				\bottomrule
			\end{tabular*}
		}
		\vspace{-2mm}
		\caption{\small Performance on single-image, bi-temporal, and short-sequence multi-image tasks (\%). \texttt{--} denotes unavailable results.}
		\label{tab:standard_rs}
		
	\end{table*}
	
	\begin{table}[tb]
		\centering
		\vspace{-2mm}
		{
			\fontsize{7.0bp}{8.6bp}\selectfont
			\setlength{\tabcolsep}{0pt}
			\renewcommand{\arraystretch}{1.18}
			
			\def\ablationlabelwidth{0.42cm}
			\def\gapAblationLabel{1pt}
			\def\leftblockwidth{3.04cm}
			\def\gapMetric{2pt}
			\def\gapLeftRight{2pt}
			\def\verticalrulewidth{0.4pt}
			
			\dimen0=\dimexpr
			\columnwidth
			-\ablationlabelwidth
			-\gapAblationLabel
			-\leftblockwidth
			-\gapMetric-\gapMetric-\gapMetric
			-\gapLeftRight-\gapLeftRight
			-\verticalrulewidth
			\relax
			\divide\dimen0 by 4
			\edef\metricwidth{\the\dimen0}
			
			\def\gapComponent{0.5pt}
			\dimen2=\dimexpr\leftblockwidth-\gapComponent-\gapComponent-\gapComponent\relax
			\divide\dimen2 by 4
			\edef\componentwidth{\the\dimen2}
			
			\begin{tabular}{
					@{}
					>{\centering\arraybackslash}m{\ablationlabelwidth}
					@{\hspace{\gapAblationLabel}}
					>{\centering\arraybackslash}m{\componentwidth}
					@{\hspace{\gapComponent}}
					>{\centering\arraybackslash}m{\componentwidth}
					@{\hspace{\gapComponent}}
					>{\centering\arraybackslash}m{\componentwidth}
					@{\hspace{\gapComponent}}
					>{\centering\arraybackslash}m{\componentwidth}
					@{\hspace{\gapLeftRight}}
					!{\vrule width \verticalrulewidth}
					@{\hspace{\gapLeftRight}}
					>{\centering\arraybackslash}m{\metricwidth}
					@{\hspace{\gapMetric}}
					>{\centering\arraybackslash}m{\metricwidth}
					@{\hspace{\gapMetric}}
					>{\centering\arraybackslash}m{\metricwidth}
					@{\hspace{\gapMetric}}
					>{\centering\arraybackslash}m{\metricwidth}
					@{}}
				\toprule
				\multirow[c]{6}{*}{\raisebox{-2.5pt}[0pt][0pt]{\rotatebox[origin=c]{90}{\scalebox{0.90}{\textbf{Component Ablation}}}}}
				& \textbf{SFT} & \makebox[0pt][c]{\textbf{Seq.IDs}} & \textbf{CoT} & \textbf{GRPO} & EvoSum & Spatial & AnomID & LogPred \\
				\cmidrule(l){2-9}
				& \xmark & \xmark & \xmark & \xmark & 40.59 & 41.21 & 29.27 & 43.74 \\
				& \cmark & \xmark & \xmark & \xmark & 60.77 & 60.27 & 29.26 & 72.49 \\
				& \cmark & \cmark & \xmark & \xmark & 70.46 & 59.43 & 72.93 & 73.82 \\
				& \cmark & \cmark & \cmark & \xmark & 69.47 & 59.08 & 64.12 & 70.81 \\
				& \cmark & \cmark & \cmark & \cmark & \textbf{74.67} & \textbf{65.55} & \textbf{76.73} & \textbf{76.85} \\
				\midrule[0.8pt]
			\end{tabular}
			
			\par\vspace{-1.15ex}
			
			\def\gapReward{0.5pt}
			\dimen2=\dimexpr\leftblockwidth-\gapReward-\gapReward\relax
			\divide\dimen2 by 3
			\edef\rewardwidth{\the\dimen2}
			
			\begin{tabular}{
					@{}
					>{\centering\arraybackslash}m{\ablationlabelwidth}
					@{\hspace{\gapAblationLabel}}
					>{\centering\arraybackslash}m{\rewardwidth}
					@{\hspace{\gapReward}}
					>{\centering\arraybackslash}m{\rewardwidth}
					@{\hspace{\gapReward}}
					>{\centering\arraybackslash}m{\rewardwidth}
					@{\hspace{\gapLeftRight}}
					!{\vrule width \verticalrulewidth}
					@{\hspace{\gapLeftRight}}
					>{\centering\arraybackslash}m{\metricwidth}
					@{\hspace{\gapMetric}}
					>{\centering\arraybackslash}m{\metricwidth}
					@{\hspace{\gapMetric}}
					>{\centering\arraybackslash}m{\metricwidth}
					@{\hspace{\gapMetric}}
					>{\centering\arraybackslash}m{\metricwidth}
					@{}}
				\multirow[c]{5}{*}{\raisebox{-3.5pt}[0pt][0pt]{\rotatebox[origin=c]{90}{\scalebox{0.90}{\textbf{Reward Ablation}}}}}
				& \raisebox{-1.5pt}{\textbf{Format}} & \raisebox{-1.5pt}{\textbf{Temporal}} & \raisebox{-1.5pt}{\textbf{Spatial}} & \raisebox{-1.5pt}{EvoSum} & \raisebox{-1.5pt}{Spatial} & \raisebox{-1.5pt}{AnomID} & \raisebox{-1.5pt}{LogPred} \\
				\cmidrule(l){2-8}
				& \xmark & \cmark & \cmark & 63.74 & 63.97 & 68.01 & \textbf{77.33} \\
				& \cmark & \xmark & \cmark & 59.83 & 64.22 & 67.02 & 76.00 \\
				& \cmark & \cmark & \xmark & 61.50 & 62.62 & 73.05 & 70.67 \\
				& \cmark & \cmark & \cmark & \textbf{74.67} & \textbf{65.55} & \textbf{76.73} & 76.85 \\
				\bottomrule
			\end{tabular}
		}
		\vspace{-2mm}
		\caption{\small Ablation results of the core components (\%). The upper block studies cumulative training components, and the lower block removes individual rewards.}
		\label{tab:ablation}
		\vspace{-5mm}
	\end{table}

	\subsection{Experimental Setups}
	
	\subsubsection{Implementation Details.}
	
	We train LongEarth-R1 on 8$\times$ NVIDIA A800 GPUs, freezing the visual encoder and tuning language-side LoRA modules with $r=128$ and $\alpha=256$ in both stages. SFT runs for two epochs with bfloat16, gradient checkpointing, cosine decay, a peak learning rate of $2\times10^{-5}$, a $0.03$ warmup ratio, and no weight decay. Inputs are limited to 8,192 tokens and interleave square-resized images with temporal prefixes and instructions. GRPO starts from the SFT checkpoint and optimizes the same LoRA modules with group-sampled responses and the proposed rewards.
	
	\subsubsection{Evaluation Metrics.}
	
	We evaluate final answers using \textit{Accuracy}, \textit{Temporal F1}, and \textit{Spatial F1}. \textit{Accuracy} applies to closed-set answers, including categorical judgments, single-frame localization, fixed spatial labels, and discrete extent levels. Temporal and free-form spatial answers use set-based \textit{F1} over predicted and reference elements. \textit{Temporal F1} operates on parsed frame indices, whereas \textit{Spatial F1} uses $\mathcal{S}=\{\text{object},\text{change},\text{region},\text{direction},\text{extent}\}$. \textit{Temporal F1} evaluates T1, T3, T8, and temporal-set variants of T7, T9, and T12; \textit{Spatial F1} evaluates free-form T4--T6 and T11. Remaining tasks use \textit{Accuracy}.
	
	\subsection{Experimental Results}
	
	\subsubsection{Performance on LongEarth-Bench.}
	
\begin{figure}[tb]
		\centering
		\includegraphics[width=\columnwidth]{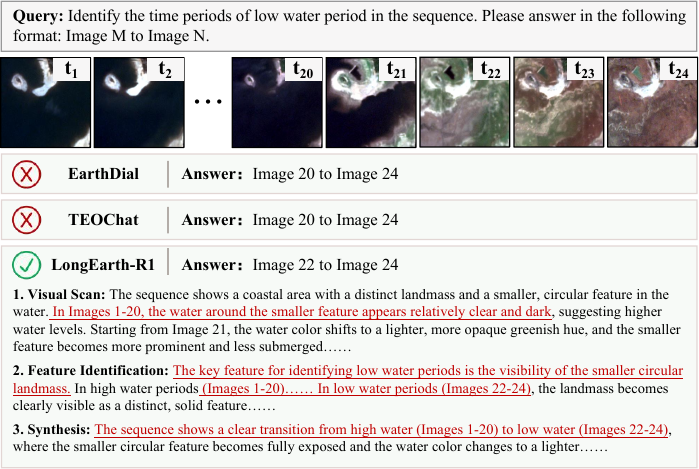} 
		\vspace{-5mm}
		\caption{\small
			Qualitative comparison on long-horizon temporal reasoning. EarthDial and TEOChat predict an incorrect interval, while LongEarth-R1 identifies the correct low-water period and grounds its answer in multi-frame visual evidence.
		}
		\label{fig:qualitative_temporal_reasoning}
		\vspace{-4mm}
	\end{figure}

	We evaluate the 12 LongEarth-Bench tasks spanning EvoSum, Spatial, AnomID, and LogPred. Table~\ref{tab:longearth_main} shows that LongEarth-R1 ranks first on all tasks, with particularly strong gains on AnomID and tasks requiring long-range temporal evidence. The consistent improvements show that sequence grounding, structured reasoning, and reward-based alignment jointly strengthen temporal understanding, spatial grounding, and prediction. Figure~\ref{fig:qualitative_temporal_reasoning} further shows that LongEarth-R1 avoids temporally misaligned intervals by grounding its answer in multi-frame evidence.

	\subsubsection{Performance on Standard Remote Sensing Tasks.}
	
	We further test whether specialization for long-term reasoning preserves general remote sensing understanding. Following the standard evaluation protocol~\cite{irvin2025teochat}, we evaluate single-image scene recognition on AID and UCM~\cite{xia2017aid,yang2010bagofvisualwords}, bi-temporal change understanding on ABCD-CD, CDVQA-QA, xBD, and S2Looking~\cite{fujita2017abcd,yuan2022cdvqa,gupta2019xbd,shen2021s2looking}, and short-sequence multi-image reasoning on Qfabric and fMoW~\cite{verma2021qfabric,christie2018fmow}. For each benchmark, LongEarth-R1 is fine-tuned on the corresponding training split and evaluated using the task-native protocol; detailed sources and task descriptions are provided in the supplementary material.
	
	Table~\ref{tab:standard_rs} shows that LongEarth-R1 achieves the best result on six of nine datasets, including all bi-temporal change-understanding benchmarks, indicating effective transfer from long-horizon supervision to conventional change analysis. Its performance remains close to the strongest specialist on the remaining benchmarks, showing that the proposed training preserves general remote sensing understanding across single-image, bi-temporal, and short-sequence settings.

	\subsubsection{Temporal Robustness Analysis.}
	
\begin{figure}[t]
		\centering
		\includegraphics[width=0.99\columnwidth]{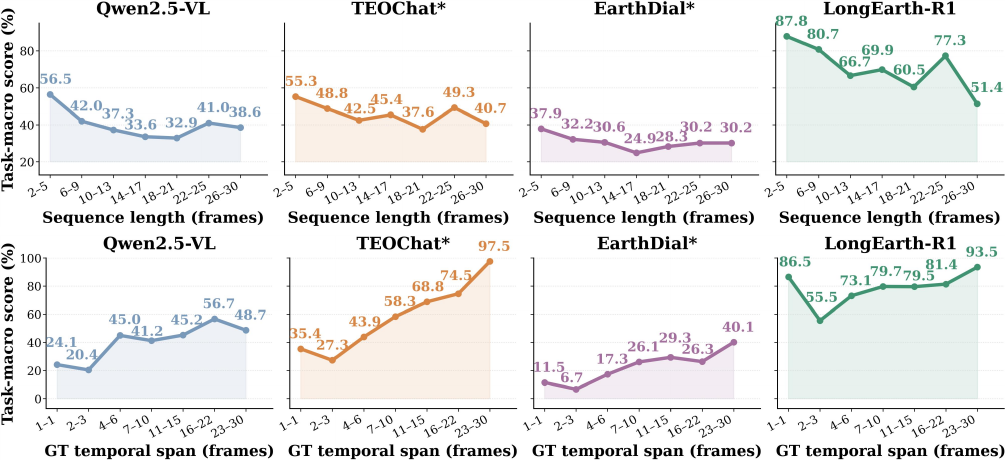}
		\vspace{-1mm}
		\caption{\small Task-macro scores by input length (top) and ground-truth evidence span (bottom); * denotes LongEarth-Bench fine-tuning.}
		\label{fig:temporal_robustness}
		\vspace{-7mm}
	\end{figure}

	Figure~\ref{fig:temporal_robustness} separates two sources of temporal difficulty: the number of input frames processed by the model and the temporal span of evidence required by the question. In the top panel, all methods degrade as input sequences become longer, reflecting the increasing need to suppress irrelevant observations and localize the informative frames. LongEarth-R1 nevertheless leads every input-length interval by 10.7--31.9\%, retaining 51.4 at 26--30 frames after reaching 87.8 on 2--5-frame inputs. This trend indicates that explicit sequence grounding and reward-based alignment improve robustness to long-context distraction, although very long inputs remain challenging.
	
	The bottom panel shows a different pattern. LongEarth-R1 achieves the best score in six of seven ground-truth evidence-span intervals, and its performance generally improves as the answer can be supported by evidence distributed across a broader temporal span. Thus, a broad evidence span is not necessarily harmful: when multiple observations provide complementary evolution cues, cross-frame reasoning can benefit from them. The only exception is the 23--30-frame interval, where TEOChat* attains a higher score.

	\subsubsection{Ablation Analysis of Core Components.}
	
	We conduct cumulative component and reward ablations across the four cognitive dimensions of LongEarth-Bench. The component study progressively adds SFT, Seq.\ IDs, R1, and GRPO, whereas the reward study removes one term at a time from the full objective.
	
	Table~\ref{tab:ablation} shows that the complete configuration, LongEarth-R1, achieves the strongest macro-average and the most balanced performance across all four dimensions. Relative to LongEarth, adding GRPO yields consistent gains, with the largest improvement on AnomID (+12.61), indicating that reward-based optimization strengthens anomaly-sensitive evidence selection and spatiotemporal reasoning. Removing any reward lowers the macro-average by 5.19--6.68\%; although removing the format reward slightly improves LogPred, the full objective remains best overall, confirming the complementarity of the three rewards.

	\section{Conclusion}
	
	Long-term remote sensing understanding requires reasoning over evolving geographic evidence rather than isolated observations. We present LongEarth-Bench, which defines this setting with four cognitive dimensions and structured reasoning supervision. We further develop LongEarth through supervised sequence grounding and LongEarth-R1 through reward-driven spatiotemporal alignment. Results improve performance across the 12 long-sequence tasks while retaining transfer to conventional remote sensing tasks. These findings support explicit modeling of temporal order, intermediate evidence, and spatial consistency for long-horizon Earth observation reasoning.

	\bibliography{reference}
\end{document}